\documentstyle[a4,epsf]{article}
\def\tm{\kern-2pt
        {\raisebox{4.2pt}{\tiny T}}\kern-1.6pt {\raisebox{4.2pt}{\tiny M}}}

\begin{document}

\begin{center}

  {\large\bf Intelligent Voice Prosthesis:}

  {\large\bf converting icons into natural language sentences}

 \vspace{.8cm}

  {\large Pascal Vaillant and Micha\"{e}l Checler}

 \vspace{.4cm}

  {\large Thomson-CSF/LCR, Cognitive Engineering Group,\\
          Advanced Human Interface Laboratory\\
          Domaine de Corbeville, F-91404 {\sc Orsay}\\
          Phone: (+33) 1 69 33 93 25, Fax: (+33) 1 69 33 08 65\\
          E-mail: \verb+vaillantp@lcr.thomson.fr+\\}

\end{center}

{\bf R\'{e}sum\'{e}~:} La {\em Proth\`{e}se Vocale Intelligente} est un
syst\`{e}me de communication qui reconstitue le sens~---~suppos\'{e}~--- d'une
s\'{e}quence peu structur\'{e}e d'ic\^{o}nes ou de symboles, et l'exprime par
des phrases en langue naturelle (fran\c{c}ais). Elle a \'{e}t\'{e}
d\'{e}velopp\'{e}e pour l'usage de personnes ne poss\'{e}dant pas la
ma\^{\i}trise du langage oral, et en particulier incapables de s'exprimer en
suivant les r\`{e}gles d'une grammaire complexe comme celle de la langue. Nous
d\'{e}crivons ici la construction d'un dic\-tion\-nai\-re s\'{e}mantique de
symboles simple et pertinent \`{a} partir des corpus de communication
ic\^{o}nique relev\'{e}s aupr\`{e}s d'enfants Infirmes Moteurs
C\'{e}r\'{e}braux (IMC). Nous expliquerons ensuite le m\'{e}canisme
d'ana\-lyse s\'{e}mantique ascendante qui permet, en d\'{e}terminant les
d\'{e}pendances entre symboles, de trouver le sens des messages de
l'utilisateur. \`{A} partir du r\'{e}sultat de cette analyse, un module de
transfert lexical choisit les mots fran\c{c}ais les mieux adapt\'{e}s pour
l'exprimer, et construit un r\'{e}seau s\'{e}mantique linguistique. Celui-ci
est ensuite hi\'{e}rarchis\'{e}, gr\^{a}ce \`{a} une Grammaire d'Arbres
Adjoints (TAG) lexicalis\'{e}e, en arbres syntaxiques de phrases
fran\c{c}aises. Enfin, nous d\'{e}crirons l'interface d'acc\`{e}s
param\'{e}trable qui a \'{e}t\'{e} d\'{e}finie pour ce syst\`{e}me.

{\bf Mots-cl\'{e}s~:} S\'{e}miotique, Analyse de D\'{e}pendances,
G\'{e}n\'{e}ration de Langue Naturelle, Handicap de Parole, Communication
Augment\'{e}e.

{\bf Abstract:} Intelligent Voice Prosthesis\footnote{The PVI ({\em
Proth\`{e}se Vocale Intelligente}) project has been funded by AGEFIPH and
Thomson-CSF. It has been the object of a CAP-HANDI contract involving the
{\em Centre de R\'{e}\'{e}ducation et R\'{e}adaptation Fonctionnelle} of
Kerpape, the LIMSI-CNRS, and Thomson-CSF's Central Research Laboratory
(Corbeville).} {\em is a communication tool which reconstructs the meaning of
an ill-structured sequence of icons or symbols, and expresses this meaning
into sentences of a Natural Language (French). It has been developed for the
use of people who cannot express themselves orally in natural language, and
further, who are not able to comply to grammatical rules such as those of
natural language. We describe how available corpora of iconic communication by
children with Cerebral Palsy has led us to implement a simple and relevant
semantic description of the symbol lexicon. We then show how a
unification-based, bottom-up semantic analysis allows the system to uncover
the meaning of the user's utterances by computing proper dependencies between
the symbols. The result of the analysis is then passed to a lexicalization
module which chooses the right words of natural language to use, and builds a
linguistic semantic network. This semantic network is then generated into
French sentences via hierarchization into trees, using a lexicalized Tree
Adjoining Grammar. Finally we describe the modular, customizable interface
which has been developed for this system.}


\newpage

%%%%%%%%%%%%%%%%%%%%%%%%%%%%%%%%%%%%%%%%%%%%%%%%%%%%%%%%%%%%%%%%%%%%%%%%%%%%%%%

\section{Introduction}
\label{introduction}

\subsection*{users' needs}

Some people are unable to speak not only because of phonatory or articulatory
reasons, but because a neurological handicap deprives them, temporarily or
permanently, of their language ability. They suffer from various types of
language difficulties, which can consist of missing words, loss of the semantic
link between a word and its meaning, inability to structure their speech, etc.

For all these people, no voice synthesis with an interface based on letter or
word selection would make up for the speech impairment. They are unable not
only to speak, but also, for various reasons, to compose a written sentence.

The users for whom the PVI system is originally designed are people with
Cerebral Palsy, in the Kerpape rehabilitation center~\cite{kn:pedelucq}. These
patients suffered from a prenatal or perinatal cerebral damage. They are to
various degrees subject to different impairments, affecting neuromotor,
articulatory, auditive, oculomotor and visual, cognitive (including linguistic)
functions. Most of them have a limited ability to progress in their mastering
of language.

The design of our system's interface, which is designed to be used by people
with {\em both} motoric and mental abilities, was carried out with constant
concern with these users' needs. We have developed various access methods for
people with specific neuromotor troubles (see~\ref{interface}).

\subsection*{iconic communication}

A solution which has already been implemented in a ``classical''
(non-electronic) way by ergotherapists for these linguistically challenged
subjects is communicating through pointing at images. The images used for
this purpose, depending on each user's abilities and knowledge, can be abstract
ideograms (for example the {\sc Bliss} alphabet~\cite{kn:bliss}) as well as
figurative pictures. Henceforth we will refer to these symbols, regardless of
their type, as {\em icons}.

During supervised communication sessions, a person used to communicating with
the speech impaired (parent, orthophonist, ergotherapist~\dots\/) goes into a
process of intelligently {\em interpreting} the sequence of icons designated,
and then formulating it back in natural language sentences. The aim of our
system is to make this process automatic and thus widely available.

The observation of practical communication situations with the patients in
Kerpape convinced us that a semantic interpretation approach was necessary for
this purpose. A simple icon-to-word translation proved to be unsufficient to
model the process: the sequences of icons used are not organized into regular
structures, but are generally arranged in an order depending on the message's
topicality.

An example of an utterance met in these iconic communication corpora will
illustrate this point:

\hspace{1cm} \makebox[6cm][l]{\tt <past-tense-indicator> boat to\_eat}

This was intended to mean ``{\em I had a meal in a boat}'' (stressing the
boat context).

No parser based on Context Free Grammars (CFG), whatever the number of rules,
can account for the different meanings of:

\hspace{1cm} \makebox[6cm][l]{\tt boat to\_eat}(``{\em I eat in a boat}'')

and:

\hspace{1cm} \makebox[6cm][l]{\tt beefsteak to\_eat}(``{\em I eat a
beefsteak}'').

The only way to cope with this type of different dependencies is to model
some of the natural expertise which allows the experienced communication
partner to assign a correct meaning to such utterances, on the basis of a
semantic interpretation.

That is why our system is based on the following processes:

\begin{itemize}

\item analysis of the semantic content of the icons used, and attribution of a
semantic role to each of them;

\item lexical choice: determining the best words to use to convey the semantic
content;

\item linguistic generation: generation of a natural language utterance from a
topicalized representation adapted to linguistic semantics.

\end{itemize}

%%%%%%%%%%%%%%%%%%%%%%%%%%%%%%%%%%%%%%%%%%%%%%%%%%%%%%%%%%%%%%%%%%%%%%%%%%%%%%%

\section{Extracting the lexicon from corpora}
\label{corpus}

The methodological choices described were supported by the analysis of corpora
of iconic communication from children with Cerebral Palsy in the Kerpape
Rehabilitation Centre.

These corpora were produced by the disabled people~\dots\/

\begin{itemize}

\item spontaneously:

\begin{itemize}

\item in a situation when they had to communicate;

\item in a situation of exercise (during ergotherapy training sessions, which
means with no communicative urge nor time limitations);

\end{itemize}

\item during supervised communication sessions:

\begin{itemize}

\item in an alternated dialogue situation, when some interlocutor could make
answers, guess meanings, complete utterances~\dots\/

\end{itemize}

\end{itemize}

The first situation will most strongly determine the design of the PVI system,
since it represents the most important requirements we are trying to
meet.\footnote{As a matter of fact, no computer program could eventually
replace the understanding skills of a human being used to communicating with
the speech impaired (parent, specialist, etc.). The system is therefore
designed, in the first place, to give the user access to an autonomous
communication device allowing him/her to talk to any person, even unprepared,
in an unspecified context.} The first corpus has thus been thoroughly
analysed, the two others giving complementary information on some elements of
the lexicon.

The corpus-based building of the lexicon guarantees the most relevant
description of the available lexical field, as it allows the system builder
(a) not to forget anything which is in the corpus, and (b) not to put anything
superfluous in the lexicon.

The analysis of possible variations on the {\em paradigmatic} dimension
(icons/symbols which can take the same place in a given context) allows the
icons to be classified into minimal classes which,
following~\cite{kn:semantique-interpretative}, we will call {\em taxemes}. An
icon representing a beefsteak and another representing a pizza will be
classified as belonging to the same taxeme as they will both tend to appear
systematically with the icon representing `{\em to eat}' in the same contexts.
These taxemes are grouped into semantic domains like ``alimentation'',
``movement'', ``game''~\dots\/

The regularities observed in the {\em syntagmatic} dimension (classes of icons
which systematically occur together in the sequences) help us build the casual
structure of predicative concepts. For example the icon representing `{\em to
eat}' will in most cases, in the corpus, go together with an icon representing
a human being or an animal, and with an icon belonging to the class which we
have identified to contain `{\em beefsteak}', `{\em pizza}',~\dots\/ This will
be expressed in the icon lexicon by giving the icon `{\em to eat}' a default
casual structure implying a first casual function which we will call
{\bf agent}, and a second casual function which we will call {\bf object}.

Each icon has its semantic content determined (a) by the taxeme it belongs to,
and (b) by elementary meaning features, the {\em specific} features, which
allow icons of the same taxeme to be distinguished.

After the analysis, the icon lexicon is given its structure:

\begin{figure}[ht]
 \begin{center}
 \epsfysize=4cm
 \mbox{\epsffile{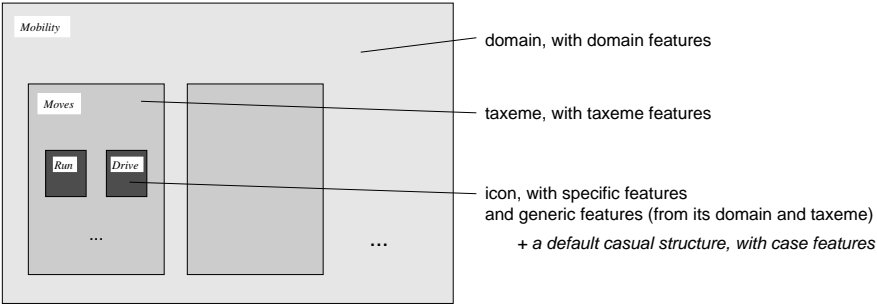}}
 \leavevmode
 \caption[lexicon]{The icon lexicon}
 \label{lexicon}
 \end{center}
\end{figure}

Without developing the theoretical issues underlying the structure of the
lexicon, we will point out that this structure naturally emerges from the
corpus analysis. In the framework of lexical
semantics~\cite{kn:semantique-interpretative}, the content is precisely based
on a description of actual use, which ensures its ability to form the basis of
interpretative processes. This approach is openly paraphrastic, and that fits
the needs of a semantic analysis system (see~\ref{analyse}).

\subsection*{What is the meaning of an icon?}

The symbols treated by PVI are represented in a structural semantic system,
where the meaning content of every icon lies in fact in the features which
distinguish it from the other icons. The elementary features used to describe
this meaning are valuated attributes (most of the time binary), which
constitute the semantic primitives of the system.

To be able to process meaning, we postulate, although few theoretical studies
on this subject support this view, that the visual icons have elementary
features which are of the same nature as linguistic semantic features. The
difference in our system between icons and words is in the possibly distinct
arrangement of these features into clusters which constitute the meaning of
icons and words. The issue of specifically visual (or non-linguistic) features
and of their role in interpretation is approached in~\cite{kn:semantique-RC}
and~\cite{kn:cavazza}.

%%%%%%%%%%%%%%%%%%%%%%%%%%%%%%%%%%%%%%%%%%%%%%%%%%%%%%%%%%%%%%%%%%%%%%%%%%%%%%%

\section{Semantic analysis}
\label{analyse}

The semantic analysis process tries to reconstruct the meaning of the sequence
of icons pointed at by the user. It builds a meaning representation of the
user's utterance, in which every icon in the sequence is attributed a semantic
role.

\begin{figure}[ht]
 \begin{center}
 \epsfysize=3cm
 \mbox{\epsffile{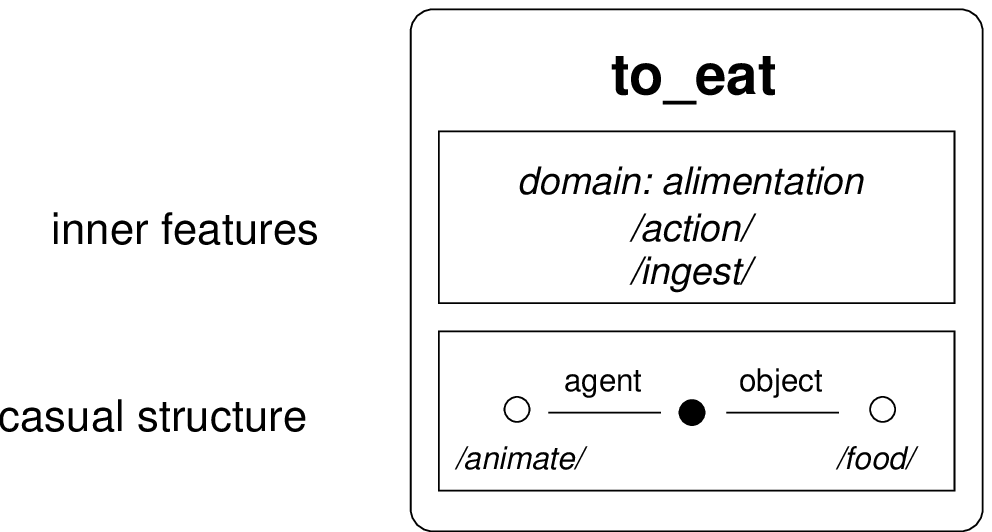}}
 \leavevmode
 \caption[icon]{An icon}
 \label{icon}
 \end{center}
\end{figure}

The information that we have {\em a priori} on the meaning of icons is (a)
inner features~---~generic or specific~---, (b) a casual structure, if the
icon has some {\em predicative} content, where case features specify the
semantic features which the functors are ``expected'' to possess (see
figure~\ref{icon}), based on observations from the corpus.

\begin{figure}[ht]
 \begin{center}
 \epsfysize=5cm
 \mbox{\epsffile{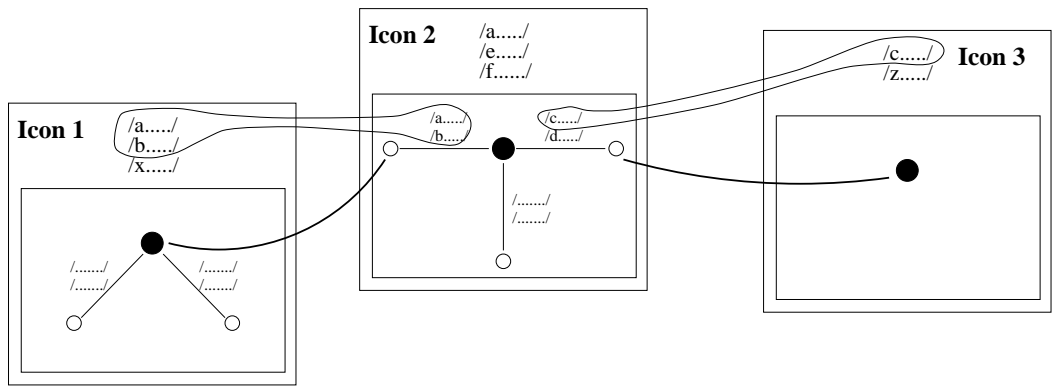}}
 \leavevmode
 \caption[unification]{Unifying the free slots of a casual structure}
 \label{unification}
 \end{center}
\end{figure}

The creation of a meaning representation then consists in assembling a
comprehensive semantic network for the utterance. This is done by assembling
the small pieces of semantic network that for every predicative icon, this
casual structure constitutes. The basic mechanism of this assembling process
is unification of the free slots of a casual structure, conditioned by the
compatibility between the semantic features (figure~\ref{unification}).

A free node in a partly instantiated network is a slot whose content is an
uninstantiated variable. If the specifications attached to the node (the case
features) are compatible with a given icon, the variable is instantiated: it
takes the icon as its value. If the icon is itself predicative, i.e. it is the
head of another partially instantiated network, this second network becomes
attached to the first. The process goes on until all possible free slots have
been unified.

The search for a solution can be described in the following way:

\begin{itemize}

  \item[1] The system scans the input sequence of icons from left to right.
  When finding a predicative icon, it looks at its casual structure and picks
  a free slot;

  \item[2] Every icon in the remainder of the sequence is then looked up,
  aiming to find one which will fill the slot, identifying it as a case filler
  of the current predicate;

  \item[3] When an acceptable solution has been found, another free slot is
  picked;

  \item[4] When an acceptable solution has been found for every slot, the
  system goes back to scan the input sequence for other predicative icons.

\end{itemize}

The notion of {\em compatibility} between semantic features, used to determine
whether an icon is an acceptable filler for a given casual slot, can be
defined in different ways depending on the selectivity expected from the PVI
system. It is a binary relation defined on two sets of semantic features.
When applied to (a) the set of ``case features'' and (b) the semantic features
of the candidate, its value characterizes the good candidates.

This binary relation may be:

\begin{itemize}

\item mere inclusion, if the semantic constraint expressed by the case
features is mandatory:

$C(a,b)$ is 1 if all features in $a$ are present in $b$ and have the same
value, 0 otherwise,

(this means that a good candidate must have {\em all} the semantic features
expected from the functor);

\item a scaled product between the two sets if more or less acceptable
solutions may be found, for example:

$C(a,b)$ is the number of features of $a$ which are present and have the same
value in $b$, divided by the total number of features in $a$,

(this means that approximate solutions are allowed).

\end{itemize}

During the analysis process, not only one solution, but many, will be explored.
This is done through the implementation in {\sc Prolog} and its backtrack
facility. It is therefore natural that we seek to find the {\em best} solution
between all the possible ones. This is the goal of ``scoring'' the solutions
with scaled semantic compatibility information such as described above.
Incomplete and ambiguous information will then be processed in the best
possible way. A similar approach has been described in~\cite{kn:al-kim}.

The result of an analysis is a semantic network expressed in a linear form.
The linear order of the network results from the processing order of the
different predicates, i.e. from the order in which they appeared in the input
sequence. It represents the topical orientation of the message.

%%%%%%%%%%%%%%%%%%%%%%%%%%%%%%%%%%%%%%%%%%%%%%%%%%%%%%%%%%%%%%%%%%%%%%%%%%%%%%%

\section{Lexical choice}
\label{transfert}

\begin{figure}[ht]
 \epsfysize=5cm
 \mbox{\epsffile{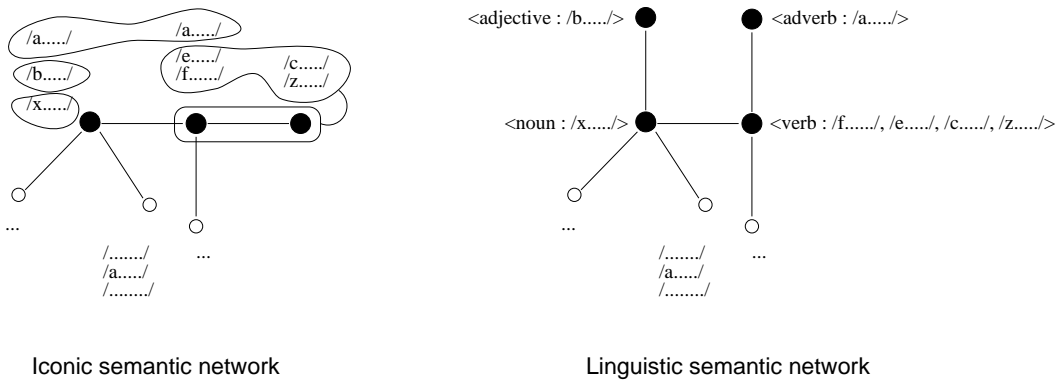}}
 \leavevmode
 \caption[lexical-choice]{The lexical choice process}
 \label{lexical-choice}
\end{figure}

Before being formulated into natural language, the semantic content of the
user's message, as resulting from the analysis, has to go through a process of
lexical choice. This process (figure~\ref{lexical-choice}) will build a
{\em linguistic} semantic network fit to be generated into natural language
words and sentences.

This lexical choice step is motivated by the observation that there is no
simple bijection between icons and words, and that their meaning content is
not necessarily isomorphic. The semantic network resulting from the analysis
of a message composed with icons is not, strictly speaking, made up of sememes
but of what we might call {\bf semioms}: clusters of semantic features which
do not necessarily match up linguistic entities.

During the lexical choice phase, clusters of semantic features of a linguistic
nature will be chosen to express those semioms in natural language in the best
possible way.

A lexical choice component has proved to be a necessary preliminary module to
any natural language generation system~\cite{kn:zock}.

The implemented mechanisms for this lexical choice component are:

\begin{itemize}

\item {\bf short-circuit:} some semantic refinement expressed explicitly with
a casual construction might be implicitly contained in the inner features of
a single word:

\makebox[3cm][l]{A $[{\rm S}_{A}]$~---~B $[{\rm S}_{B}]$}
\makebox[1cm][l]{$\rightarrow$}
L $[{\rm S}_{A} \bigcup {\rm S}_{B}]$

\item {\bf derivation:} some icons with too ``rich'' a content for natural
language might have to be expressed with more than one word:

\makebox[3cm][l]{I $[{\rm S}_{A} \bigcup {\rm S}_{B}]$}
\makebox[1cm][l]{$\rightarrow$}
A $[{\rm S}_{A}]$~--- B $[{\rm S}_{B}]$.

\end{itemize}

These mechanisms have already been explored for automatic translation studies,
the problem of lexical choice being also an important issue in this
field\footnote{translaters are familiar with this type of content
redistribution, every natural language having its unique way of expressing
meaning, like the German phrase `{\em \"{u}ber den Flu\ss{} schwimmen}' being
expressed rather in French by `{\em traverser la rivi\`{e}re \`{a} la nage}'.}.


\section{Generation}
\label{generation}

The last phase is the generation of a natural language sentence conveying the
meaning of the linguistic semantic network. This operation is based on the
principle that every sememe may be expressed through a small number of lexemes
(in many cases one~---~sometimes two or three, depending on the syntactic
function, e.g. `work' $[${\em noun}$]$ vs. `to work' $[${\em verb}$]$), each
of which is in its turn lexicalised through a certain number of morphemes (for
inflexion).

The semantic structure of casual relations linking a sememe to its casual
fillers is itself expressed in natural language through a morpho-syntactic
structure which native speakers of a language will identify. Among the
mechanisms that natural languages have developed in this purpose, French uses
chiefly the following three:

\begin{itemize}

\item word order (e.g. $<$subject$>$~$<$verb$>$~$<$object$>$~\dots\/);

\item inflexion (plural of nouns, conjugation of verbs~\dots\/);

\item functional morphemes (e.g. ``{\em \`{a}}'' ({\em at, to}), ``{\em de}''
({\em of, from}), ``{\em sur}'' ({\em on})~\dots\/).

\end{itemize}

Our lexicon stores, for every entry, elementary syntactic trees representing
possible phrase constructions. Each of these elementary syntactic trees
specifies the following information:

\begin{itemize}

\item the lexeme corresponding to the sememe;

\item the morphosyntactic structure expressing its casual structure.

\end{itemize}

This is a {\em lexicalized grammar}.

\begin{figure}[ht]
 \begin{center}
 \epsfysize=6cm
 \mbox{\epsffile{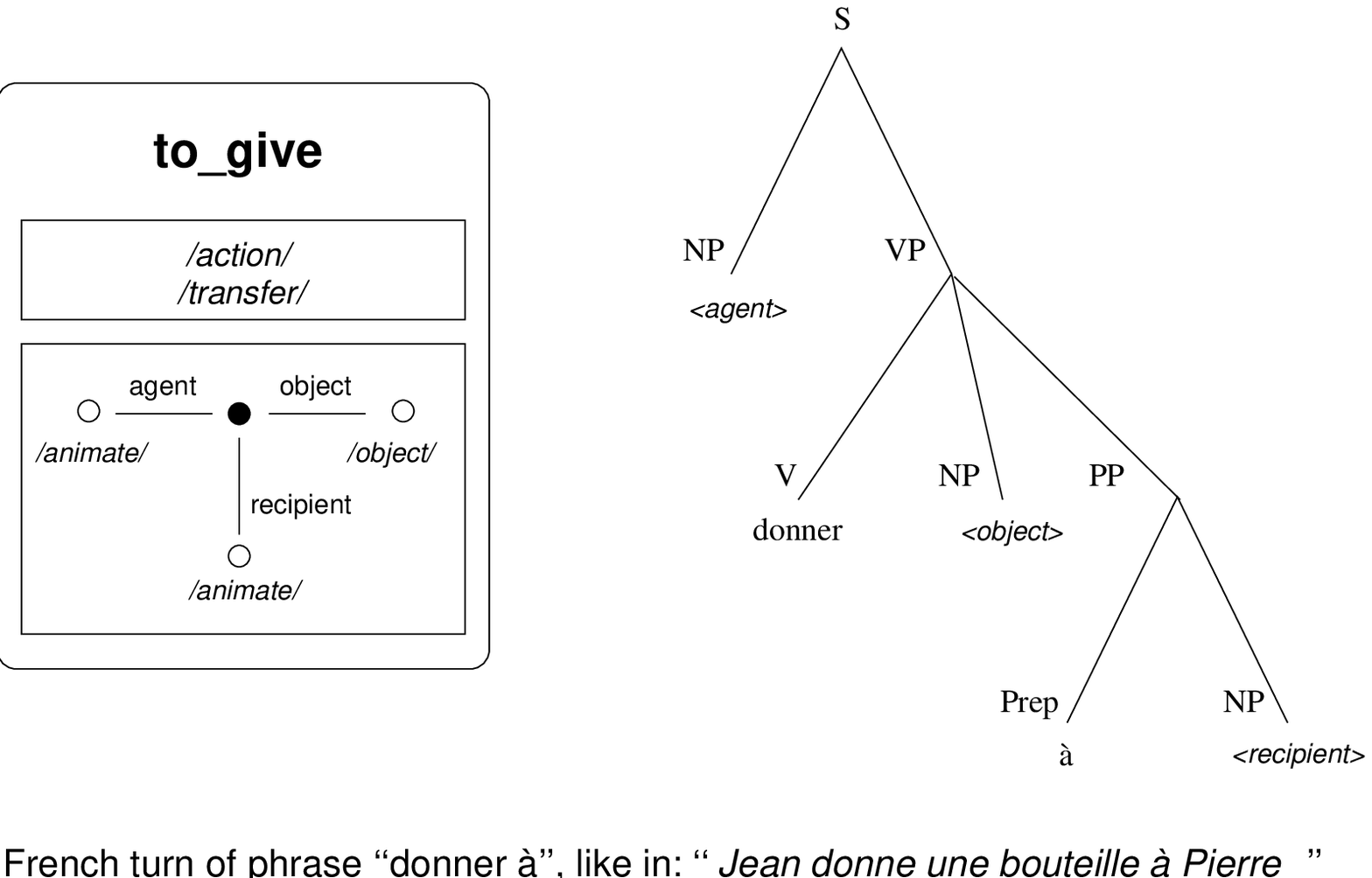}}
 \leavevmode
 \caption[case-structure]{A semantic casual structure and a possible
                          morphosyntactic expression}
 \label{case-structure}
 \end{center}
\end{figure}

The elementary trees contain the information necessary to specify the three
mechanisms listed above. The word order is reflected in the tree structure;
the functional morphemes, if any, are terminal nodes of the elementary tree;
the flexion is given by constraints propagated from upper to lower nodes in
the tree (unification features). The example in figure~\ref{case-structure}
illustrates this.

In our system, the generation of the sentence corresponding to the semantic
network is done during a scan over the network, considering predicates in turn
in the topical order, following the semantic links. It corresponds to a one
shot run over a spanning tree of the network.

For every node (sememe) met in the network, a corresponding elementary tree is
selected. Elementary trees are assembled using the following operations:

\begin{itemize}

\item {\bf substitution}, for the ``compulsory'' functors\\
(a branch is ready for them in the tree; e.g. {\em agent}, {\em object} and
{\em recipient} in the tree figure~\ref{case-structure});

\item {\bf adjunction}, for the ``optional'' functors\\
(like, for example, if we had a locative complement in the example above),

or for a new predicative sememe for which an already generated sememe acts as
functor\\
(like, for example, a qualifying adjective).

\end{itemize}

\begin{figure}[ht]
 \begin{center}
 \epsfysize=5cm
 \mbox{\epsffile{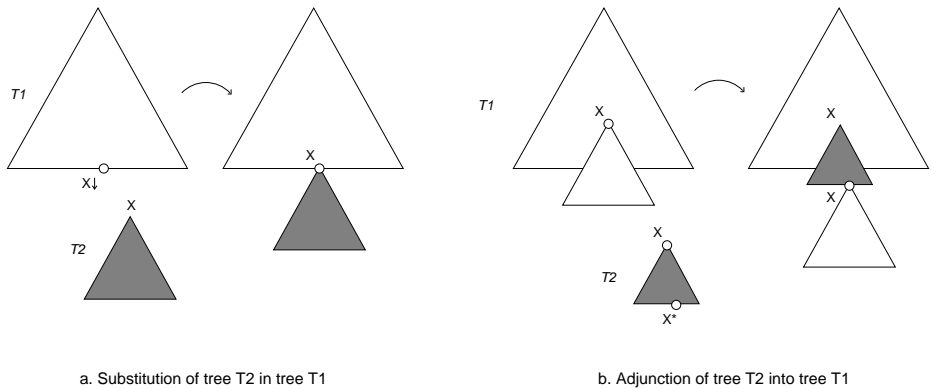}}
 \leavevmode
 \caption[tag]{The operations on trees in TAGs}
 \label{tag}
 \end{center}
\end{figure}

These two operations on trees, substitution and adjunction (see
figure~\ref{tag}), define a Tree Adjoining Grammar (TAG). Such grammars have
been introduced by~\cite{kn:joshi}, and a French adaptation has been proposed
by~\cite{kn:abeille}.

When the system comes to a new predicative icon which cannot be generated
through an adjunction to the sentence tree currently being built, a new
sentence tree is generated. A network can thus be generated into more than
one tree.

The output sentences are the list of inflected morphological forms of the
terminal nodes of those trees. These morphological forms are found in a
morphological lexicon.

The sentences are eventually vocalized by a text-to-speech voice synthesis
device.

%%%%%%%%%%%%%%%%%%%%%%%%%%%%%%%%%%%%%%%%%%%%%%%%%%%%%%%%%%%%%%%%%%%%%%%%%%%%%%%

\section{Interface}
\label{interface}

Some potential users of the PVI system have neuromotor troubles which make the
use of classical, widespread interfaces difficult for them. One of the goals
of the system is to provide them with an adapted interface.

A Human-Computer Interface (HCI) consists of a set of material devices and
logical routines allowing the person and the computer to exchange information.
The PVI system is to be installed and run on a Macintosh\tm{} type computer,
which offers in its standard operating system a graphical window manager
interface and supports many input/output devices. Some disabled people may not
use very well the most common input devices: keyboard, mouse, trackball or
joystick. Some more specific devices, such as a tongue contactor or ultrasonic
``headphones'' (a head-commanded pointing device) may be used in some cases.
The simplest device might consist of a mere push button.

These users are compelled to use simplified hardware devices. The information
supplied to the machine is all the poorer. The support of these devices thus
demands from the interface software specific methods to assist the user in
his/her interaction with the machine.

\subsection*{Architecture}

In the case we are interested in, we expect the interface to allow the user to
point at a sequence of symbols, and to be able to synthetize an oral French
sentence. Given the variability of possible neuromotor impairments challenging
some users of PVI, and the diversity of devices able to be used, it is vital
to design an open and flexible system. The interface for PVI is in nearly
every detail customizable.

The graphical interface is a segmented window displaying icons. It deals with
four data types: text, pictures, sounds and moving pictures. The internal
representation of an icon contains data of one or more of these types, which
means that an icon can be as much a sound icon as a visual icon. For example,
the `cat' icon displays the drawing of a cat, the string ``{\em chat}'', and a
meowing sound.

Apart from their perceptive content, the icons may be of two types: symbols or
actions. The symbols are transmitted to the semantic analyser, whereas the
actions directly command the parameters of the interface (change window,
louder sound, etc.).

The windows are used to group homogeneous icons. All the static aspects
(position, size, color of icons and windows, sound volume, etc.) and all the
dynamic aspects (pointing method used, editing commands, etc.) of the
interface are customizable.

\subsection*{Dynamics}

What are the methods that may be used to point at the icons composing the
message?

In a first case, the user has sufficient motor abilities to use a pointing
device commanding a graphic cursor (like a mouse). (S)he may then click on the
selected icon, or else leave the cursor unmoved for a minimum time to activate
an implicit selection.

In a second case, the user is able to use a keyboard and (s)he may access a
large number of keys. Every key will then correspond to an icon on the screen,
possibly on a special keypad. This is direct access designation.

In the last case, the user's mobility is reduced and (s)he can only use a
binary information device (like a push button). The system has to make up for
the missing information by a motion automaton: since the user cannot move the
cursor, the cursor moves from one square to another in the window, and the
user just has to validate the selected icon when the cursor is in front of it.

There are more or less sophisticated motion automata:

\begin{itemize}

\item the cursor is displayed on every square in the window in turn;

\item the cursor is a window-wide line and is displayed on each line in turn
in the window. The user stops it when it is positioned on the line where the
selected icon is. Then a cursor moves from one square to another along the
line.

\item the cursor is a surface which can encircle a region of the screen. The
window is divided into groups of squares. The cursor is displayed in turn on
each of these groups of squares. When the user has selected one group, the
cursor may continue to move, within it, on subgroups of squares, or on the
squares themselves.

\end{itemize}

For example, if the window is composed of 32 squares arranged on 4 lines and
8 columns, designating an icon might take from 1 to 32 moves with the first
method, from 2 (1+1) to 12 (4+8) moves with the second one, and~---~assuming
that the selected group of squares is divided in two at each step~---, will
always take 5 moves with the third method.

%%%%%%%%%%%%%%%%%%%%%%%%%%%%%%%%%%%%%%%%%%%%%%%%%%%%%%%%%%%%%%%%%%%%%%%%%%%%%%%

\section{Conclusion}
\label{conclusion}

The PVI system is a fully-implemented system, although as yet limited to a
small semantic domain. It carries out the entire processing chain for
converting messages from one sign system to another. For the needs of our
application, we were led to develop a semantic interpretation mechanism to
understand the icon sequences. We also have developed a generation module
which implements the operations on trees in Tree Adjoining Grammars; these
operations constitute a good model to express semantic unification in natural
language.

PVI has been designed to have a reasonable robustness inside its domain. For
field test, the system has entered a preliminary validation phase in the
Kerpape Rehabilitation Centre. It has received encouraging comments regarding
its modularity and flexibility, which are important features for disabled
users.

Future work will be dedicated to a subtler analysis of the relation between
the semantic contents of two different sign systems. Context analysis should
also lead to a better processing of ambiguities and reference.

%%%%%%%%%%%%%%%%%%%%%%%%%%%%%%%%%%%%%%%%%%%%%%%%%%%%%%%%%%%%%%%%%%%%%%%%%%%%%%%

\section*{Acknowledgements}

We would like to thank J.-P. Departe (Kerpape Centre) and M. Zock (LIMSI) for
their valuable advice on the requirements and advancement of PVI, A. Werts
(Thomson) for his support and advice about leading the project, and M. Cavazza
(Thomson) for valuable discussions during the project, and detailed comments
on this paper. We also owe thanks for useful comments to M. Zurfluh, X.
Pouteau and J. Fowler.

\end{document}